\pdfoutput=1

\documentclass[11pt]{article}

\usepackage[]{acl}

\usepackage{times}
\usepackage{latexsym}

\usepackage[T1]{fontenc}

\usepackage[utf8]{inputenc}
\usepackage{footmisc}

\usepackage{microtype}

\usepackage{inconsolata}

\usepackage{graphicx}
\usepackage{amsfonts} 
\usepackage{booktabs} 
\usepackage{multirow} 
\usepackage{subfigure} 
\usepackage{makecell}

\title{What Kinds of Tokens Benefit from Distant Text?\\An Analysis on Long Context Language Modeling}

\author{
    Yutong Hu$^{1,2}$, 
    Quzhe Huang$^{1,2}$, 
    Kangcheng Luo$^{3}$, 
    \textbf{Yansong Feng}$^{1}$\thanks{\;\;Corresponding author.}~~\\ 
    $^1$Wangxuan Institute of Computer Technology, Peking University, China \\ 
    $^2$School of Intelligence Science and Technology, Peking University \\
    $^3$School of Electronics Engineering and Computer Science, Peking University, China \\ 
    {\tt \{huyutong,huangquzhe,fengyansong\}}
     {\tt @pku.edu.cn} \\
     {\tt luokangcheng@stu.pku.edu.cn}\\
}

\begin{document}
\maketitle
\begin{abstract}

As the context length that large language models can handle continues to increase, these models demonstrate an enhanced ability to utilize distant information for tasks such as language modeling. This capability contrasts with human reading and writing habits, where it is uncommon to remember and use particularly distant information, except in cases of foreshadowing. In this paper, we aim to explore which kinds of words benefit more from long contexts in language models.
By analyzing the changes in token probabilities with increasing context length, we find that content words (e.g., nouns, adjectives) and the initial tokens of words benefit the most. Frequent patterns in the context (N-grams) also significantly impact predictions. Additionally, the model's prior knowledge plays a crucial role in influencing predictions, especially for rare tokens. We also observe that language models become more confident with longer contexts, resulting in sharper probability distributions. This overconfidence may contribute to the increasing probabilities of tokens with distant contextual information.
We hope that our analysis will help the community better understand long-text language modeling and contribute to the design of more reliable long-context models.


\end{abstract}

\section{Introduction}


%

Many studies have expanded the context window of Large Language Models (LLMs) to process longer inputs, leading to the development of long-context LLMs~\cite{Zhu2023PoSEEC,chen2023longlora,ding2023longnet,yarn}. To explore whether a model can effectively process extremely long texts, the language modeling task is frequently evaluated. This task is straightforward to calculate and can be extended to inputs of any length. Language modeling tests the accuracy of predicting the next word based on the previously input text. Previous research has found that as the context length increases, the model's performance in language modeling improves, indicating that the model can utilize more distant information.

When examining models' performance on long-context language modeling tasks, we discover an interesting phenomenon: as the context length increases, the model's language modeling ability continues to improve, even when the input text is very long, such as 32k tokens. This phenomenon does not align with human writing and reading habits. Except for cases involving foreshadowing, it is rare for people to recall particularly distant information in everyday writing or reading, let alone use such distant information to assist in their subsequent writing. This may be because humans, when solving problems, are limited by their working memory and find it difficult to remember too much information simultaneously~\cite{WMcapacity}. 

In long-context language modeling tasks, we are curious (1)\textit{Does only a small number of words benefit from ultra-long contexts, just like humans perform foreshadowing?}
If not, (2)\textit{what kinds of tokens benefit from the additional distant text in long-context LLMs' language modeling?}

To answer this question, we compare the tokens' probabilities among different context lengths. We find that, when context length increases, there are more tokens whose probability scores increase while fewer whose scores decrease. This indicates a large part of tokens can benefit from the additional distant text, different from human habits 
that only a few tokens are related to distant foreshadowings. So we wonder, what kinds of tokens benefit from the additional distant text in long-context LLMs' language modeling? 
To answer this question, we plan to analyze it from three aspects: the characteristics of the \textbf{words} themselves, the \textbf{context} in which the tokens appear, and the \textbf{priors} inherent in large language models.

We analyzed the changes in the predicted probabilities of different tokens as context length increases, using Longlora, Yarn, and Yi as representative long contexts. The greater the increase in a token's predicted probability with increasing context length, the more we consider that token's prediction to be influenced by the long context. By comparing the changes in predicted probabilities for different types of tokens, we derived the following conclusions:

(1). From the perspective of the words themselves, we found that \textbf{content words such as nouns and adjectives benefit more from longer contexts.} On the other hand, in current large models, words are often split into several tokens. We discovered that the first token of a word is more influenced by the length of the context, while the predicted probabilities of subsequent subwords remain relatively stable. (2). From the perspective of the context, \textbf{frequent patterns in long texts, i.e., N-grams, have a significant impact on the prediction of the next word.} We observed that if the additional context frequently provides an N-gram containing the token to be predicted, the model can predict this token with a higher probability. (3). We also found that \textbf{the model's prior has an influence, though possibly not as significant as the impact of the context.} We observed that if a token appears very frequently in the pre-training corpus, meaning the model has a strong grasp of this token, its predicted probability is less likely to be affected by the length of the context. However, for tokens that are rare in the training corpus, their predicted probabilities are more likely to change with the context length. The above conclusions hold only when the impact from the frequent patterns does not change as the context length increases. Otherwise, the probability changes are mainly influenced by the context and have no significant correlation with the model's prior.

We also find that, apart from the tokens aforementioned benefit from the distant text, tokens that are incorrectly predicted by LLMs also show a higher average probability.
We speculate that long-context LLMs are more confident as the context length increases. Specifically, with longer input, the probability distribution predicted by LLMs becomes sharper, i.e., the max probability becomes larger, regardless of whether the model correctly predicts the token. Therefore, overconfidence may be one of the reasons that perplexity continuously decreases as the context length increases.

To summarize, the contributions are as follows:
\begin{enumerate}
\item [1)] We find the inconsistency between humans' behavior and LLMs' performance on long-context language modeling. Humans are unlikely to use distant information for their subsequent writing, while for long-context LLMs, a large number of tokens can benefit from distant text to have a higher probability.
\item [2)] We analyze what kinds of tokens are more likely to benefit from distant context from aspects of the characteristics of the \textbf{words} themselves, the \textbf{context} where the tokens appear, and the \textbf{priors} inherent in LLMs, providing insights for a better understanding of long-context LLMs.
\item [3)] We find the growing confidence of LLMs as the context length increases, which may be one explanation for the continuously decreasing perplexity. Thus, researchers should carefully consider the impact of overconfidence to design more reliable long-context LLMs.
\end{enumerate}


\section{Preliminary}

\label{sec:perplexity_def}

\subsection{Perplexity} 
Perplexity is widely used to evaluate language models.
As it can be calculated on any text without length limitation, perplexity is widely used for LLMs' long context processing ability evaluation~\cite{press2021train,longnet,Zhu2023PoSEEC}. Following~\citet{press2021train}, we use the sliding window evaluation of perplexity in evaluating long-context LLMs. Specifically, given $M$ documents $\{D_1, ..., D_M\}$ and context length $K$, long-context LLMs predict each token $x^t_i$ in each document $D_t$ based on previous $K$ tokens when $i\geq K$, otherwise, based on all its previous tokens:
\begin{equation}
    \label{eq:prob}
    p_K(x^{t}_{i}) = P(x^{t}_{i}|x^{t}_{0}, ..., x^{t}_{i-1}), i \in [0, K)
\end{equation}
\begin{equation}
    \label{eq:prob}
    p_K(x^{t}_{i}) = P(x^{t}_{i}|x^{t}_{i-K}, ..., x^{t}_{i-1}), i \in [K, |D_t|)
\end{equation}
We illustrate the sliding window evaluation on the left of Figure~\ref{fig:chunk}.
After obtaining all tokens' probabilities in $D_t$, perplexity (PPL) is calculated in the same way as~\citet{chen2023longlora}:
\begin{equation}
    \label{eq:token_perplexity}
    \bar{p}_K(x^{t}_i)=-log(p_K(x^{t}_i)), i \in [0, |D_t|)
\end{equation}
\begin{equation}
    \label{eq:perplexity}
    PPL_K = \frac{1}{M}\sum\limits_{t=1}^{M}(\frac{1}{|D_t|}\sum\limits_{i=0}^{|D_t|-1}\bar{p}_K(x^t_i))
\end{equation}
where we denote $\bar{p}_K(x^t_i)$ as {\bf token-perplexity} in our paper. According to the definition of perplexity, the lower perplexity indicates a better language modeling ability\footnote{Please refer to~\citet{press2021train} for more details about sliding window perplexity evaluation.}. 

\subsection{Perplexity Decreases as Context Length increases}
Recently, many LLMs have been proposed with the ability to handle extremely long context. For example, GPT-4~\cite{openai2023gpt4} has a context window of 128k, while Yi~\cite{ai2024yi} can even process 200k tokens. 

When using perplexity to evaluate the language modeling ability of long-context LLMs, a vast majority of previous works~\cite{press2021train,Zhu2023PoSEEC,chen2023longlora,peng2023YaRN} unanimously report that perplexity decreases as the context length $K$ increases. This indicates the model can understand the entire document better~\cite{zhang2023survey}. Such a phenomenon still exists when $K$ scales up from 32k to 64k~\cite{chen2023longlora,peng2023YaRN}. However, this phenomenon does not align with human writing habits. With limited working memory capabilities, it is difficult for humans to remember all information that is too far away~\cite{WMcapacity}. Therefore, except for foreshadowings, humans may seldom refer to distant information when writing subsequent words. Words in the document are more likely to be related to local text. So we wonder if the models perform like humans so that the decrease in perplexity comes from a few tokens, such as tokens related to foreshadowing. If not, this indicates that not only tokens related to foreshadowings, but also some other tokens written based on local information by humans can benefit from the distant text by long-context LLMs, which provokes us to wonder, what kinds of these tokens are?

\section{Experimental Setup}
To answer the aforementioned questions, we need to have a detailed analysis regarding the token-perplexity of each token in documents predicted by long-context LLMs.

\paragraph{Models}
We select three representative long-context LLMs for experiments, which use different methods to scale up the context window to more than 100k. 1) Yi-6B-200K~\citep{ai2024yi}, which adjusts the base frequency of position embedding Rotary Position Embeddings (RoPE)~\cite{su2023roformer} for context window extension. 2) YaRN-7B-128K~\citep{yarn}, which extends RoPE by interpolating frequencies unevenly and keeping high frequencies intact. 3) LongLoRA-7B-100K~\citep{chen2023longlora}, which proposes shifted sparse attention (S2-Attn) to approximate long context learning while retaining the original attention architecture during inference.

\begin{figure}[t]
    \center
    \includegraphics[width=0.5\textwidth]{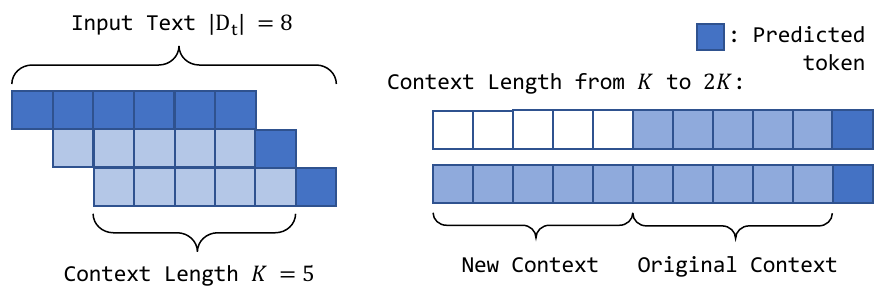}
    \caption{Left part: an illustration for sliding window method of perplexity calculation. Right part: an illustration of {\it original context} and {\it new context}.}
    \label{fig:chunk}
\end{figure}


\paragraph{Dataset}
Following~\cite{chen2023longlora}, we calculate perplexity on PG-19~\citep{rae2019compressive}, a language modeling benchmark, including a set of books derived from the Project Gutenberg books project\footnote{https://www.gutenberg.org}. Due to the computation resources limitation, we randomly sample 6 books from the test set of PG-19 as our test corpus for experiments. The lengths of selected books are all larger than 64k after tokenization.

\paragraph{Setup}
First of all, the sliding window evaluation method mentioned in~\ref{sec:perplexity_def} is time-consuming because every time LLMs predict a token, the input text needs to be re-encoded. Following~\citet{alibi}, we make predictions on $S$ tokens instead of one every inference pass to improve the computational efficiency. Please refer to Appendix~\ref{app:stride} for more details about {\it stride} $S$.


Secondly, we use the change of token-perplexity when context length increases to determine whether the token benefits from the additional text. Specifically, we compare the token-perplexity between the context length of $K$ and $2K$ in our experiment. The illustration is shown in the right part of Figure~\ref{fig:chunk}. For simplicity, we name the input context when context length is $K$ as "{\it original context}" $[x^t_{i-K+1}, ..., x^t_{i-1}]$, while naming the newly added context when context length is extended to $2K$ as "{\it new context}" $[x^t_{i-2K+1}, ..., x^t_{i-K}]$ in our paper. When the token-perplexity decreases, we say that the token can benefit from the {\it new context}.

\section{Most Tokens' Token-perplexity Decrease}
\begin{table}[t]
    \centering
    \scriptsize
    \setlength\tabcolsep{4.0pt}
    \begin{tabular}{c|c|c|c|c|c|c}
        \toprule[2pt]
        \multicolumn{2}{c|}{{\bf $K$}} & {\bf 2k} & {\bf 4k} & {\bf 8k} & {\bf 16k} & {\bf 32k}\\
        \midrule
        \multirow{3}{*}{\makecell[c]{Decrease\\Ratio}} & Yi & 52.6\% & 52.0\% & 49.2\% &	45.8\% & 39.8\% \\
        & YaRN & 55.0\% &	53.0\% & 51.4\% & 49.0\% & 44.5\% \\
        & LongLoRA & 54.6\% &	54.1\% & 54.7\%	& 50.1\% & 45.3\% \\
        \midrule
        \multirow{3}{*}{\makecell[c]{Increase\\Ratio}} & Yi & 43.7\%	 & 43.8\% & 41.8\% & 39.8\%	& 35.4\% \\
        & YaRN & 45.0\% & 45.3\% & 45.2\% & 44.2\% & 41.8\% \\
        & LongLoRA & 45.3\% & 45.8\%	& 45.1\% & 43.1\% & 41.0\% \\
        \bottomrule[2pt]
    \end{tabular}
    
    \caption{Ratio of tokens with token-perplexity decrease and increase.}
    \label{tab:dec_ratio}
\end{table}

    

\begin{table}[t]
    \centering
    \small
    \setlength\tabcolsep{4.0pt}
    \begin{tabular}{c|c|c|c|c|c|c}
        \toprule[2pt]
        {\bf $K$} & {\bf 2k} & {\bf 4k} & {\bf 8k} & {\bf 16k} & {\bf 32k} & {\bf 64k} \\
        \midrule
        Yi & 2.105 &  2.078 & 2.055 & 2.037 & 2.024 & 2.015 \\
        YaRN & 1.944 &1.918 & 1.896 & 1.880 & 1.867 & 1.859 \\
        LongLoRA & 2.084 & 2.051 & 2.022 & 2.001 & 1.985 & 1.976 \\
        \bottomrule[2pt]
    \end{tabular}
    
    \caption{Perplexity on test corpus.}
    \label{tab:ppl}
\end{table}

We first verify that the phenomenon that perplexity decreases as $K$ increases still exists in our experimental setting. In Table~\ref{tab:ppl}, we report the perplexity of three long-context LLMs with $K$ ranging from 2k to 64k. The results show that perplexity consistently decreases as $K$ increases. 

Therefore, we want to figure out whether the decrease in perplexity comes from a uniform decrease in most token-perplexity or from a drastic decrease in some token-perplexity. We count the proportion of tokens with a decreased token-perplexity to the total number of tokens when the context length is extended from $K$ to $2K$. 

The experimental results are shown in Table~\ref{tab:dec_ratio}. The sum of each decrease ratio and its counterpart is less than 100\% since some token-perplexity remains unchanged. 
In almost all settings, the token-perplexity decrease ratios are larger than 40\%, and consistently surpass the increase ratio as well. Therefore, we can conclude that the decrease in perplexity comes from a decrease in the token-perplexity of a large number of tokens, which contradicts human habits that most tokens are only written with local information. 

\section{What Tokens Benefit from Distant Text}
To investigate this issue, we conduct the analysis from three perspectives: characteristics of the \textbf{words} themselves, the \textbf{context} where the tokens appear, and \textbf{priors} inherent in long-context LLMs.

\subsection{Properties of Words}


\begin{figure*}[t]
    \center
    \includegraphics[width=1\textwidth]{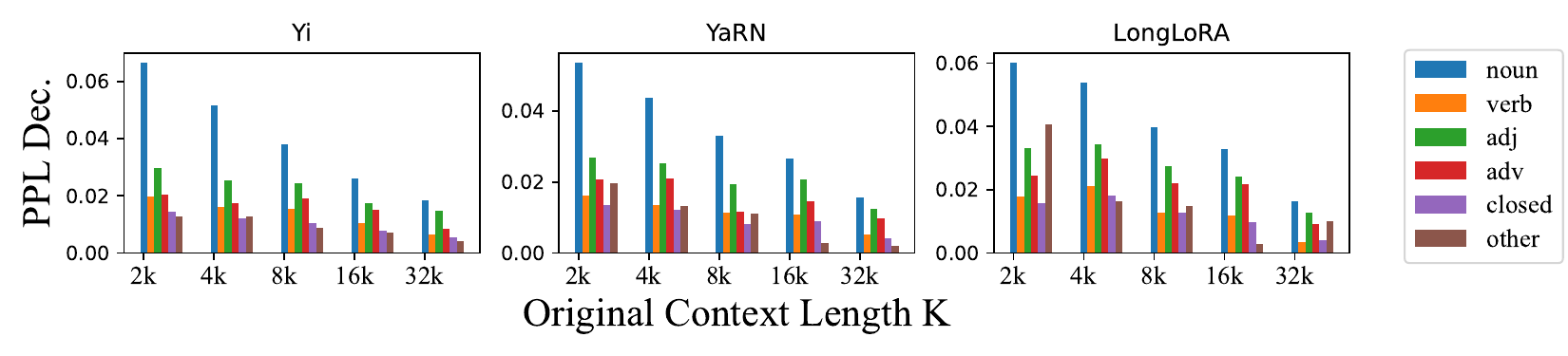}
    \caption{The average token-perplexity decrement in each class of POS tags.}
    \label{fig:pos}
\end{figure*}

\paragraph{Lexical Property.}
\label{sec:pos}
Function words, such as adpositions that play a grammatical role, and content words (semantically richer words), such as proper nouns that contain more information, are two major groups of parts of speech (POS). \citet{BELL200992} mentions that function words are much more frequent and predictable than content words. As content words contain more information while function words rarely change form or meaning in different contexts, we wonder, provided with additional text, whether content words can benefit more than function words as content words may be more related to global information. Specifically,  we want to investigate the correlation between the types of POS and the decreased value of token-perplexity in longer context length.

To determine the POS of each token $x_i$, we first find its corresponding word $w_i$ in the document before tokenization. Then we use Stanford CoreNLP~\cite{manning-etal-2014-stanford} to obtain the POS of $w_i$, and we treat it as the POS of token $x_i$ as well. According to the definition of the POS tag set~\footnote{\url{https://www.ling.upenn.edu/courses/Fall\_2003/ling001/penn\_treebank\_pos.html}}\footnote{\label{fn:pos}\url{https://web.stanford.edu/~jurafsky/slp3/slides/8_POSNER_intro_May_6_2021.pdf}}, we divide them into six main classes: {\it noun}, {\it verb}, {\it adj}, {\it adv}, {\it closed} (such as adposition, particle, pronoun, etc.) and {\it other} (not a word, such as punctuation, symbols, etc.). Among the six classes, {\it noun}, {\it verb}, {\it adj}, and {\it adv} are called "open words" (content words), which usually contain important information, while {\it closed} means "closed words" (function words), which are used to stitch "open words" together.

To compare the token-perplexity between the context length of $K$ and $2K$, we calculate the {\bf token-perplexity decrement} of each token $x_i$ as:
\begin{equation}
    \label{eq:prob_change}
    \Delta \bar{p}_{K}^{2K}(x_i) = - (\bar{p}_{2K}(x_i) - \bar{p}_{K}(x_i))
\end{equation} 
The larger $\Delta \bar{p}_{K}^{2K}(x_i)$, the larger decrement of token $x_i$'s token-perplexity. For brevity, we abbreviate it as $\Delta \bar{p}(x_i)$

We further calculate the average token-perplexity decrement in each class of POS tags on the test corpus, with {\it original context} length $K$ ranging from 2k to 32k. The results are shown in Figure~\ref{fig:pos}. We can observe that "{\it noun}" and "{\it adj}" deliver the largest token-perplexity decrement in all context length $K$ while tokens in "{\it closed}" and "{\it other}" show the minimal decrease in most cases. These results indicate the additional information in {\it new context} benefits "{\it noun}" and "{\it adj}" most. 

\begin{figure*}[t]
    \center
    \includegraphics[width=1\textwidth]{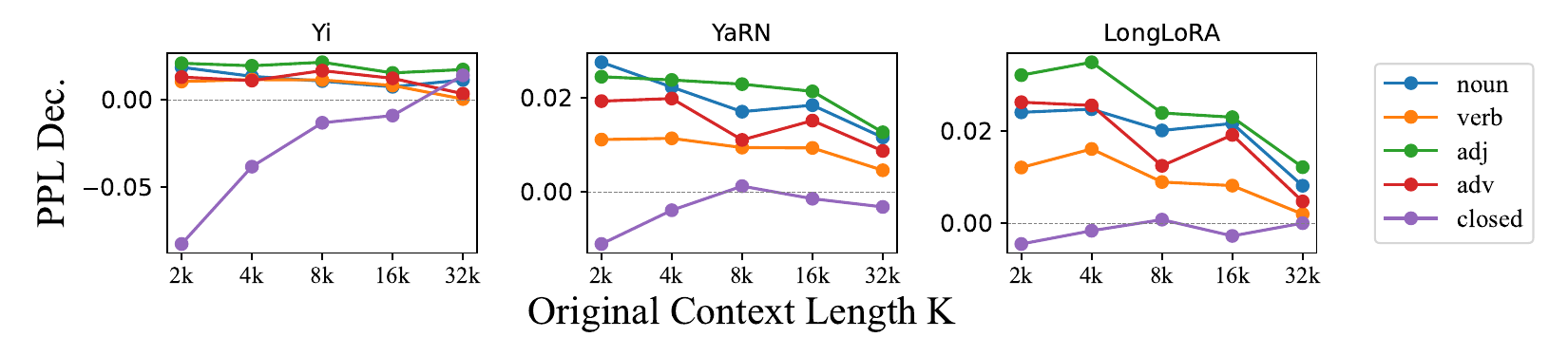}
    \caption{$\Delta D$ of each class of POS tags.}
    \label{fig:subword_idx}
\end{figure*}

\subsubsection{Structures inside Words}
Apart from examining tokens that are easily influenced by {\it new context} from a lexical perspective, we can also consider the structure inside the word itself. Specifically, after tokenization, a word $w_j$ may be split into multiple tokens $[x_{w_j, 0}, ..., x_{w_j, n}]$. The first token $x_{w_j, 0}$ can only be predicted based on previous words while subsequent tokens can be predicted by leveraging previous tokens in the same word. Therefore, we wonder whether LLMs need more information to infer the starting token $x_{w_j, 0}$, that is, whether the first token of a word will benefit more from the {\it new context}, resulting in a larger decrement in token-perplexity.

To evaluate the relationship between the tokens' position in words, we compare the token-perplexity decrement $\Delta \bar{p}(x_i)$ of the first tokens $x_{w_j, 0}$ and the later tokens $x_{w_j, i(i\neq 0)}$ in words. Specifically, we separate all tokens in the test corpus into two sets: {\bf Fir}(st) and {\bf Lat}(ter) according to whether the token is the first token of its original word. Then we calculate the average token-perplexity decrement of each set and compare the decrement value between two sets. The difference between the token-perplexity decrement of the two sets is computed as:
\begin{equation}
    \Delta D = \frac{1}{|Fir|}\sum\limits_{i\in Fir}\Delta \bar{p}(x_i) - \frac{1}{|Lat|}\sum\limits_{i\in Lat}\Delta \bar{p}(x_i)
\end{equation}

Considering POS classes are an important factor of token-perplexity decrement, we calculate $\Delta D$ for each class of POS. The results are shown in Figure~\ref{fig:subword_idx}. The results of class "{\it other}" are not reported because all words in "{\it other}" can not be tokenized into multiple tokens. Except for "closed", $\Delta D$ of all classes of POS is larger than 0, which demonstrates that long-context LLMs deliver a larger token-perplexity decrement in the first tokens with additional text. We can also observe that, the first tokens of POS classes "{\it noun}" and "{\it adj}" decrease more than other POS classes in most cases. Therefore, the first token of open words, especially for POS classes "{\it noun}" and "{\it adj}", can benefit most from the {\it new context}.

\subsection{Influence of Context}
\begin{table}[t]
    \centering
    \small
    \setlength\tabcolsep{5.0pt}
    \begin{tabular}{c|c|c|c|c|c}
        \toprule[2pt]
        {\bf $K$} & {\bf 2k} & {\bf 4k} & {\bf 8k} & {\bf 16k} & {\bf 32k} \\
        \midrule
        Yi & 0.530 & 0.347 & 0.417 & 0.410 & 0.356 \\
        YaRN & 0.342 & 0.455 & 0.461 & 0.349 & 0.345 \\
        LongLoRA & 0.380 & 0.408 & 0.344 & 0.340 & 0.307 \\
        \bottomrule[2pt]
    \end{tabular}
    
    \caption{Correlation coefficients between the token-perplexity decrement $\Delta \bar{p}$ and the N-gram's new occurrence ratio $\Delta \mathcal{N}$. All reported correlation coefficients have p-values < 0.005.}
    \label{tab:ngram}
    \vspace{-3mm}
\end{table}

\paragraph{Effect of N-gram's Occurrence.}
If a phrase frequently appears in the context, LLMs are more likely to pay attention to this phrase. Therefore, when given the first few words of the phrase, LLMs may attend to the phrase which appears multiple times in the previous context, and predict the latter words accurately. From this perspective, we want to figure out whether the more an N-gram appears in the input text, the more possible the long-context LLMs refer to the N-gram for prediction, i.e., whether there is a correlation between the token $x_i$'s token-perplexity and the number of the token's N-gram $g_i=[x_{i-N+1}, ..., x_i]$ occurrences.

We first count the number of N-gram $g^t_i$'s occurrences in the {\it original context} $[x^t_{i-K+1}, ..., x^t_{i-1}]$ and the {\it new context} $[x^t_{i-2K+1}, ..., x^t_{i-K}]$ of document $D_t$, which we denote as $\mathcal{N}^t_{ori, i}$ and $\mathcal{N}^t_{new, i}$ respectively. The ratio between $\mathcal{N}^t_{ori, i}$ and $\mathcal{N}^t_{new, i}$ represents how the N-gram's occurrence frequency changes when adding {\it new context}, which we denote as new occurrence ratio:
\begin{equation}
    \Delta \mathcal{N}_{K}^{2K}(g^t_i) = \frac{\mathcal{N}^t_{new, i}+1}{\mathcal{N}^t_{ori, i}+1}, i \in [2K-1, |D_t|)
\end{equation}
Then we calculate the average token-perplexity decrement $\Delta \bar{p}_{K}^{2K}$ and the average N-gram's new occurrence ratio $\Delta \mathcal{N}_{K}^{2K}$ over all documents:
\begin{equation}
    \Delta \bar{p}_{K}^{2K} = \frac{1}{M}\sum\limits_{t=1}^{M}(\frac{1}{|D_t|}\sum\limits_{i=2K-1}^{|D_t|-1}\bar{p}_{K}^{2K}(x^t_i))
\end{equation}
\begin{equation}
    \Delta \mathcal{N}_{K}^{2K} = \frac{1}{M}\sum\limits_{t=1}^{M}(\frac{1}{|D_t|}\sum\limits_{i=2K-1}^{|D_t|-1}\mathcal{N}_{K}^{2K}(g^t_i))
\end{equation}
which we abbreviate as $\Delta \bar{p}$ and $\Delta \mathcal{N}$ for brevity.

To figure out the relationship between the token-perplexity decrement $\Delta \bar{p}$ and the N-gram's new occurrence ratio $\Delta \mathcal{N}$, we adopt a widely used metric, Spearman's rank correlation coefficient, for analysis. Specifically, Spearman's rank correlation coefficient is computed over $\Delta \bar{p}(x^t_i)$ and $\Delta \mathcal{N}(g^t_i)$ of all tokens $x^t_i$ in the test corpus. Here we show the result of $N$=5 and $K$ ranges from 2k to 32k, and we will discuss the effect of $N$ later. 

As the results shown in Table~\ref{tab:ngram},  
there is a strong correlation coefficient in every experimental setting. For example, with $K=32k$, Yi delivers a correlation coefficient of 0.356. When $K=2k$, the correlation coefficient is even higher, up to 0.53, demonstrating the positive correlation between the token-perplexity decrement $\Delta \bar{p}$ and N-gram's new occurrence ratio $\Delta \mathcal{N}$. The larger the new occurrence ratio of a token's N-gram, the more its token-perplexity decreases. Therefore, tokens with a higher frequency of N-gram in {\it new context} can benefit more from the additional long text.

\begin{figure}[t]
    \center
    \includegraphics[width=0.38\textwidth]{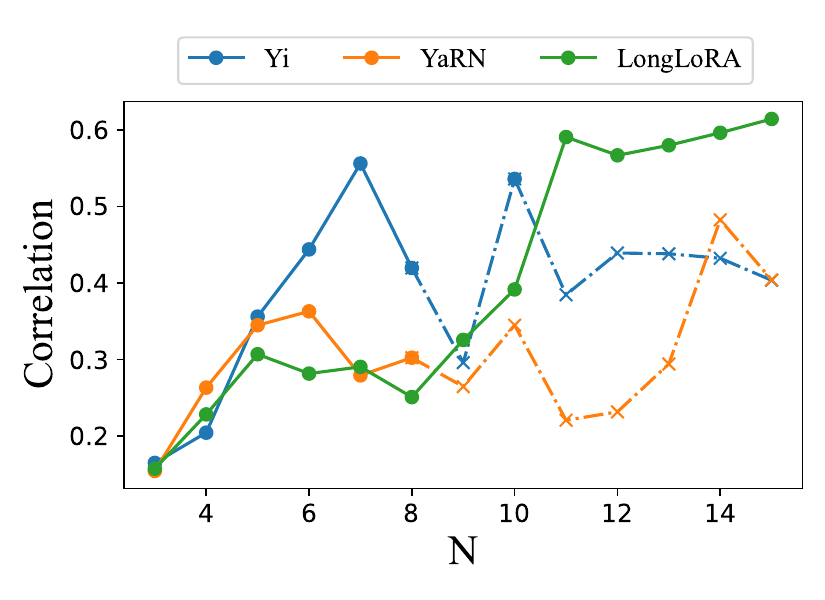}
    \caption{Correlation coefficients between the token-perplexity decrement $\Delta \bar{p}$ and the N-gram's new occurrence ratio $\Delta \mathcal{N}$ under different values of N.}
    \label{fig:ngram}
\end{figure}

\paragraph{Effect of N.}
We further analyze the effect of $N$ on the correlation between $\Delta \bar{p}$ and $\Delta \mathcal{N}$. We fix $K$ to 32k and calculate the Spearman's rank correlation coefficient with $N$ ranging from 3 to 20, as shown in Figure~\ref{fig:ngram}. Circle markers represent correlation coefficients with p-value < 0.005 while x markers represent p-value $\geq$ 0.005, which indicates there is no correlation.

There are always correlations between $\Delta \bar{p}$ and $\Delta \mathcal{N}$ when $N$ ranges from 4 to 8. In LongLoRA, we can see the correlation become stronger with the increase of $N$, indicating under the same $\Delta \mathcal{N}$, a longer N-gram may more easily affect its corresponding token's prediction. Considering the impressive in-context learning ability~\cite{brown2020language} of LLMs, it may be one possible explanation for the strong correlation between $\Delta \bar{p}$ and $\Delta \mathcal{N}$ that LLMs tend to learn the N-grams frequently appearing in the input text. 

\subsection{Priors in Long-context LLMs}
Priors in long-context LLMs may affect their performance in language modeling. For example, as pre-training is a crucial part of LLM training, the composition of pre-training data can greatly affect the LLMs' performance in downstream tasks. So, in the language modeling task, is the decrease in token-perplexity also affected by pre-training data? Specifically, if a token appears more frequently in the pre-training data and the LLM is fully familiar with this token, will the LLM be more sensitive, i.e., less affected by changes in the context, when predicting this token? From this perspective, we will explore the relationship between the frequency of tokens appearing in pre-training data and the change in token-perplexity.

\paragraph{Token's Frequency Calculation.} As the data used to
pre-train LLMs are rarely disclosed, we use RedPajama~\cite{together2023redpajama}, a fully open-source reproduction of LLaMA~\cite{touvron2023llama1}, as a proxy for calculating the tokens' frequency in LLMs' pretraining dataset. We randomly sample 20G tokens from RedPajama, where the proportion of each subset is determined according to the pretraining data sampling proportion mentioned in LLaMA. We calculate the frequency of each token $fr(x_i)$ in the sampled dataset to approximate the token frequency in the pre-training data.

\begin{table}[t]
    \centering
    \small
    \begin{tabular}{c|c|c|c}
        \toprule[2pt]
        $\Delta \mathcal{N}$& $(0,1)$ & $\{1\}$ & $(1, \infty)$  \\
        \midrule
        Yi & 0.049 & 0.463* & 0.086  \\
        YaRN & 0.014 & 0.330* & 0.107 \\
        LongLoRA & 0.047 & 0.385* & 0.111 \\
        \bottomrule[2pt]
    \end{tabular}
    
    \caption{Correlation coefficients between the token-perplexity decrement $\Delta \bar{p}$ and the token frequencies $fr$ under different N-gram's new occurrence ratio $\Delta \mathcal{N}$. $K$=32k, $N$=5. *: p-value < 0.005.}
    \label{tab:fix_ngram}
\end{table}

\paragraph{Correlation between $fr$ and $\Delta \bar{p}$ under Different $\Delta \mathcal{N}$.} 
To evaluate the relationship between the token frequency and its token-perplexity change, we calculate the correlation between the token frequency and the degree of changes in token-perplexity. Note that, unlike the token-perplexity decrement $\Delta \bar{p}$ in the previous experiments, this experiment examines the degree of token-perplexity {\bf changes} affected by the token frequency, which is calculated as:
\begin{equation}
    \delta \bar{p}_{K}^{2K}(x_i) = abs(\bar{p}_{2K}(x_i) - \bar{p}_{K}(x_i))
\end{equation}
which we denoted as $\delta \bar{p}(x_i)$ for brevity.

Considering that the new occurrence ratio of a token's n-gram $\Delta \mathcal{N}(g^t_i)$ will also affect its token-perplexity change $\delta \bar{p}(x_i)$, we classified tokens into 4 groups based on $\Delta \mathcal{N}(g^t_i)$. This guarantees that the $\Delta \mathcal{N}(g^t_i)$ of the token $x_i$ within the same group are close to each other, thus mitigating the impact of context information on token-perplexity changes. We group the tokens according to the following rules: Group A: $\{x_i|\Delta \mathcal{N}<1\}$, Group B: $\{\Delta \mathcal{N}=1\}$, Group C: $\{x_i|\Delta \mathcal{N}>1\}$. For each group, we calculate the correlation coefficient between $\delta \bar{p}$ and $fr$. 

The results are shown in Table~\ref{tab:fix_ngram}. Only when $\Delta \mathcal{N}=1$, i.e., the frequency of N-gram $g_i$ does not change when adding new context, there is a strong correlation between $fr(x_i)$ and $\delta \bar{p}(x_i)$, such as correlation coefficient of 0.463 in Yi. In other cases, there is no correlation as their p-value > 0.005. The results indicate that the context information dominates the influence on the token-perplexity changes than tokens' frequency in the pretraining dataset. The long-context LLMs are more likely to use what they learned during the pretraining stage to predict the current token $x_i$ when the N-gram $g_i$'s frequency does not change in the context.

\section{Why Perplexity Decreases} 

\begin{figure*}[t]
    \center
    \includegraphics[width=0.95\textwidth]{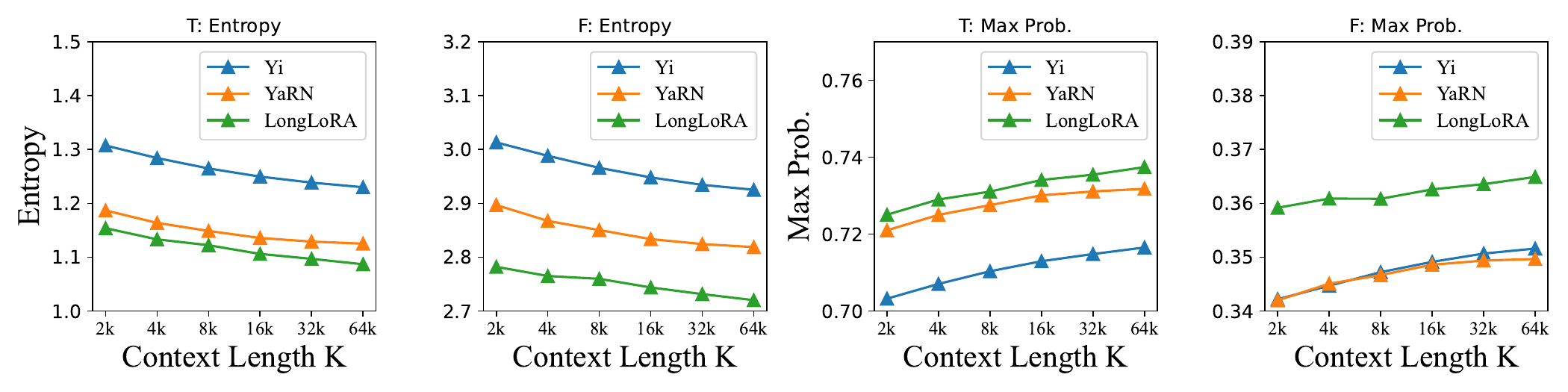}
    \caption{The entropy $E_K$ and the max probability $MP$ of groups {\bf T} and {\bf F} respectively.}
    \label{fig:entropy}
\end{figure*}

We also find that, apart from the tokens aforementioned benefit from the distant text, tokens that are incorrectly predicted by LLMs also show a higher average probability. So we wonder, whether long-context LLMs exhibit overconfidence when the context length increases.
Specifically, we explore the probability distribution $\mathbb{P}$ output by the model when predicting each token. If the model becomes more confident, no matter whether or not the model correctly predicts the token $x_i$, the probability distribution $\mathbb{P}_{K,i}$ will become sharper, i.e., the entropy of $\mathbb{P}_{K,i}$ decreases and the $max(\mathbb{P}_{K,i})$ increases.

Therefore, we calculate the entropy $E_{K,i}=entropy(\mathbb{P}_{K,i})$ and the max probability $MP_{K,i}=max(\mathbb{P}_{K,i})$ of each token's probability distribution $\mathbb{P}_{K,i}$. Specifically, we split the tokens in the whole test corpus into two groups, {\bf T} and {\bf F}, based on whether the tokens are correctly predicted ({\bf T}) by the model or not ({\bf F}). Given context length $K$, for each group, we calculate the average entropy $E_K$ and the average max probability $MP_K$ of all tokens in the group following the equation:
\begin{equation}
    E_K^{\bf T} = \frac{1}{|{\bf T}|}\sum_{i\in {\bf T}}E^t_{K,i}
\end{equation}
\begin{equation}
    MP_K^{\bf T} = \frac{1}{|{\bf T}|}\sum_{i\in {\bf T}}MP^t_{K,i}
\end{equation}
Here we use group {\bf T} as an example, $E_K^{\bf F}$ and $MP_K^{\bf F}$ of group {\bf F} are calculated in the same way. 


Figure~\ref{fig:entropy} shows that, as the context length $K$ increases, there are consistent trends between group {\bf T} and {\bf F} in both $E_K$ and $MP_K$. Especially in group {\bf F}, where LLMs make incorrect predictions, $E_K^{\bf F}$ decreases and $MP_K^{\bf F}$ increases, indicating the longer inputs lead to sharper probability distributions. Such a phenomenon shows that long-context LLMs are more confident with longer inputs. 

Therefore, for group {\bf T}, the $MP_K^{\bf T}$'s increase may be partly due to the sharper probability distributions from the more confident long-context LLMs. Note that, all tokens in {\bf T} satisfy $argmax(\mathbb{P}^{\bf T}_{K,i})=i$, i.e., $MP_{K,i}=p_K(x_i)$. According to the definition of perplexity in Equation~\ref{eq:token_perplexity}, an increase in $MP_{K,i}$ will lead to a decrease in token-perplexity, which demonstrates that the increasing confidence of the long-context LLMs may be one of the reasons for the perplexity decrease.

\section{Related Work}
\paragraph{Long-Context LLMs.}

Extensive studies have aimed to scale up the context window of Large Language Models (LLMs) to handle long-context inputs~\cite{openai2023gpt4,claude2,chen2023extending,chen2023clex,xiong2023effective, ding2023longnet, chen2023longlora}. For example, GPT-4~\cite{openai2023gpt4} has a 128k context window, and Yi~\cite{ai2024yi} supports a context window of 200k.

Some long-context LLMs use the length extrapolation approach in Transformers~\cite{transformer}, which is trained on short sequences while inferring on long sequences, to handle long text~\cite{alibi, xpos,rope}. While some other research, such as Position Interpolation~\cite{chen2023extending}, NTK-aware position embeddings~\cite{blocntkaware}, and YaRN~\cite{yarn}, propose positional interpolation methods for long text processing.

Besides, LLMs like LongLoRA~\cite{longlora} and LongNet~\cite{ding2023longnet} focus on the efficient attention calculation. Retrieval-based approaches~\cite{focused-transformer,longmem,borgeaud2022improving}, recurrent transformers~\cite{bulatov2022recurrent,staroverov2024recurrent} and prompt compression~\cite{longllmlingua} are also effective methods for context window extension.

\paragraph{Long-context LLMs Evaluation.}
Multiple benchmarks have been proposed for long-context LLMs evaluation. ZeroSCROLLS~\cite{zeroscrolls} is a zero-shot benchmark containing
ten natural language tasks. L-Eval~\cite{leval} encompasses 18
realistic natural language tasks, such as QA, summarization, and math. Similarly, LongBench~\cite{bai2023longbench} incorporates 21 tasks of four categories. InfiniteBench~\cite{inftybench} is proposed with average data length surpassing 100K tokens. RULER~\cite{hsieh2024ruler} proposes synthetic tasks of four categories, aiming to provide flexibility to control the context lengths and task complexities.

Apart from downstream tasks such as QA, summarization and retrieval, language modeling is also widely used for long-context LLMs evaluation~\cite{longlora,yarn,ai2024yi,longnet,men2024base}, which use Perplexity (PPL) as the evaluation metric to access LLMs' long text language modeling ability.

\section{Conclusion}
Different from human habits, a great number of tokens can benefit from additional distant text in long-context language modeling of LLMs. Specifically, content words and the starting token of a word benefit most from the long text. Patterns' frequency (N-gram) also plays an important role in token predictions. Besides, tokens of high frequency in the pre-training dataset show less sensitivity to the extension of the text. Furthermore, we observe that the overconfidence of long-context LLMs when the context length increases may be one possible reason for the perplexity decrease. We hope our analysis can provide insights for a better understanding of long-context LLMs and help the community design more reliable long-context LLMs.

\section*{Limitations}
Due to the computational resources limitation, we can only conduct experiments with context length $K$ ranging from 2k to 64k. We get out-of-memory when $K$=128k. Besides, it is worthy to further investigate why some tokens continuously benefit from the additional text even if the text is extremely far away. We will leave it as our future work.



\bibliography{custom}

\clearpage

\appendix

\section{Sliding Window Evaluation of Perplexity}
\label{app:stride}
\begin{figure}[h]
    \center
    \includegraphics[width=0.4\textwidth]{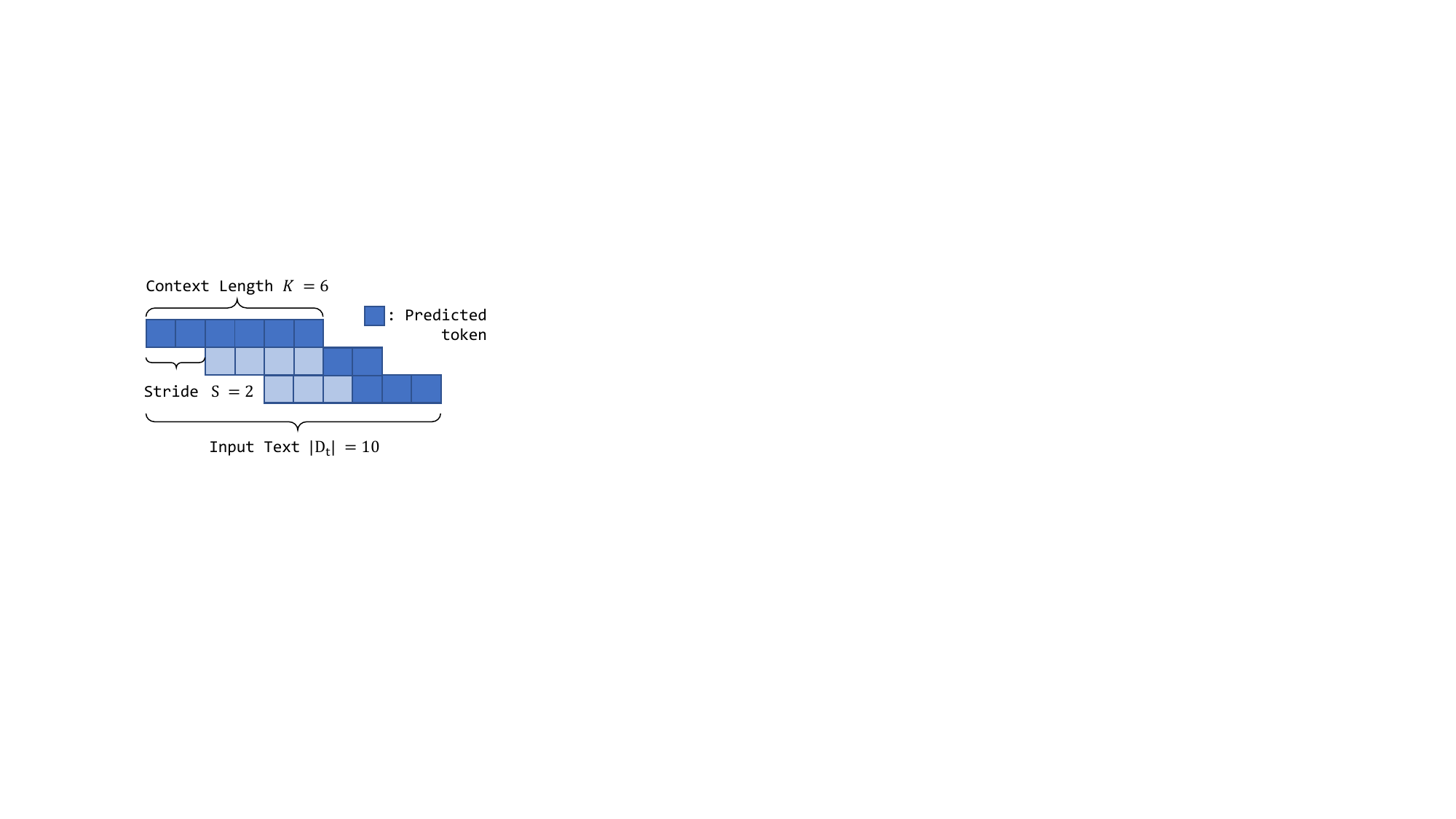}
    \caption{An illustration for sliding window method with stride $S$=2.}
    \label{fig:chunk_stride}
\end{figure}

\begin{table}[h]
    \centering
    \begin{tabular}{c|c|c|c|c|c|c}
        \toprule[2pt]
        $K$ & 2k & 4k & 8k & 16k & 32k & 64k \\
        \midrule
        $S$ & 10 & 25 & 50 & 100 & 200 & 400 \\
        \bottomrule[2pt]
    \end{tabular}
    
    \caption{Context length $K$ and its corresponding stride $S$.}
    \label{tab:k_and_s}
\end{table}
Here, we will briefly introduce the sliding window evaluation~\cite{press2021train} of perplexity in evaluating long-context LLMs. Given $M$ documents $\{D_1, ..., D_M\}$, each $D_t$ is split into chunks with stride $S$ and the length of each chunk is $K$, which is also denoted as {\bf context length} $K$. In each chunk, long-context LLMs predict token $x_i$ based on previous tokens in the chunk $C_q$, and output the probability of token $x_i$:
\begin{equation}
    \label{eq:prob}
    p_K(x^{t}_{q, i}) = P(x^{t}_{q, i}|x^{t}_{q, 0}, ..., x^{t}_{q, i-1}), i \in [0, K)
\end{equation}

As illustrated in Figure~\ref{fig:chunk_stride}, we obtain the probabilities of all tokens in the first chunk. Meanwhile, for other chunks, we only obtain the probability of the last $S$ tokens. This way, we can finally get all tokens' probabilities in $D_t$. Then perplexity (PPL) is calculated as Equation~\ref{eq:perplexity}. The calculation method we mentioned in Section~\ref{sec:perplexity_def} is the case where $S$=1.

Note that, except for the first chunk, probabilities of tokens in other chunks are only recorded when $i \in [K-S, K)$. For fair comparison among all tokens' token-perplexity, we need to ensure every token is predicted based on a similar length of input text. Therefore, in our experiment, we set $K>>S$ to ensure all tokens are predicted by long-context LLMs based on nearly the same number of previous tokens. The values of $K$ and $S$ are shown in Table~\ref{tab:k_and_s}.

\end{document}